%% file: main.tex
\documentclass[10pt,twocolumn,letterpaper]{article}

\usepackage[pagenumbers]{cvpr} 


\definecolor{cvprblue}{rgb}{0.21,0.49,0.74}
\usepackage[pagebackref,breaklinks,colorlinks,allcolors=cvprblue]{hyperref}

\usepackage{url}
\usepackage{booktabs}
\usepackage{multirow}
\usepackage{algorithmic}
\usepackage{xcolor,colortbl}
\usepackage{graphicx} 
\usepackage{enumitem}
\usepackage{amsmath}
\usepackage{amssymb} 
\usepackage{bm}     
\usepackage[normalem]{ulem}
\usepackage{cancel}

\usepackage[linesnumbered,ruled,vlined]{algorithm2e}
\usepackage{xcolor}

\usepackage{listings}

\newcommand{\newshortname}{LLaVA-1.5}
\newcommand{\ourshortname}{Panther}
\usepackage{xcolor}
\usepackage[symbol]{footmisc}
\definecolor{commentcolor}{RGB}{110,154,155}   
 
\def\paperID{} 
\def\confName{CVPR}
\def\confYear{2025}

\title{\raisebox{-0.5em}{\includegraphics[height=1.8em]{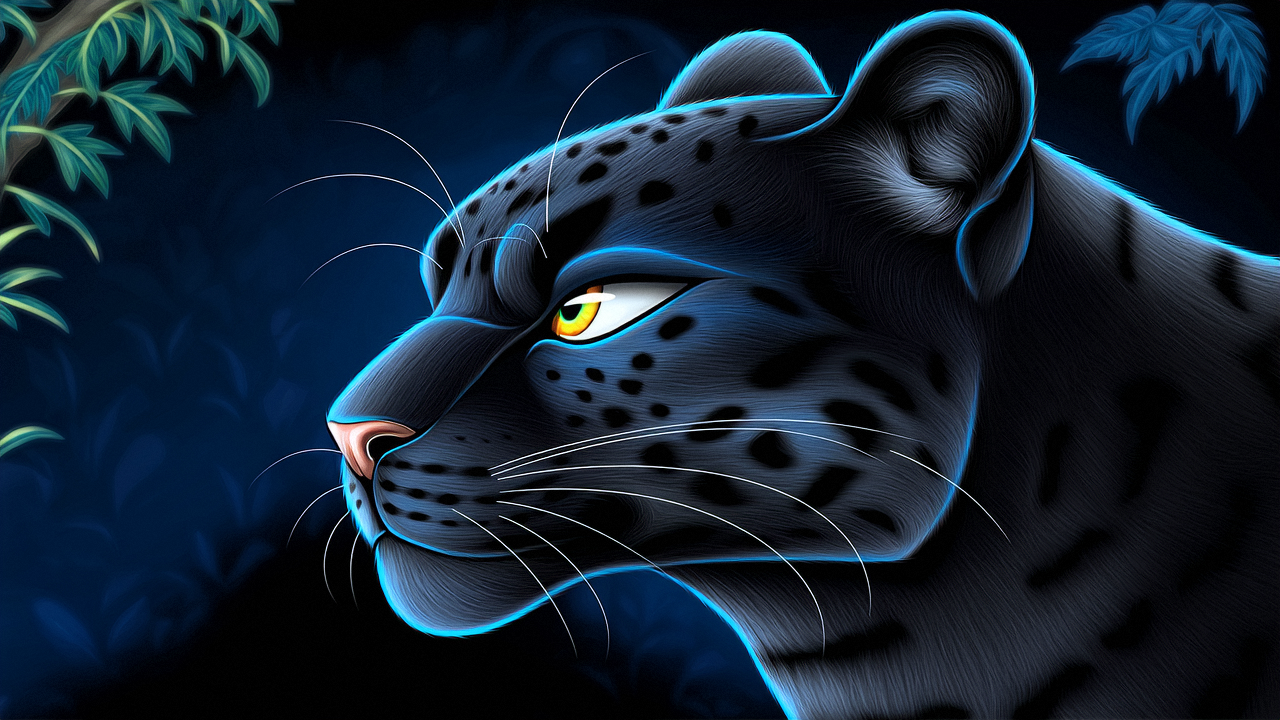}}
Panther: Illuminate the Sight of Multimodal LLMs  with \\ Instruction-Guided Visual Prompts}

\author{Honglin Li$^{1,2}$, Yuting Gao$^{3}$, Chenglu Zhu$^{2}$, Jindong Chen$^{3}$, Ming Yang$^{3}$, Lin Yang$^{2}$ \\
$^{1}$ Zhejiang University $^{2}$ Westlake University    $^{3}$ Ant Group\\
}

\begin{document}
\maketitle
\footnotetext{lihonglin@westlake.edu.cn}
\input{sec/0_abstract}    
\input{sec/1_intro}
\input{sec/2_related_work}
\input{sec/3_method}

\input{sec/4_experiments}

\section{Conclusions}
This paper addresses the visual capability shortcomings in multimodal large language models (MLLMs) and proposes a new method, Panther, to tackle the specific \textit{Amblyopia} issue in encoder-decoder architectures. 
The Panther improves MLLMs' ability to locate and focus on instruction-specific details within images. It consists of three key components: Panther-VE, Panther-Bridge, and Panther-Decoder. Panther-VE incorporates user instruction in the vision encoding process, enhancing the extraction of relevant visual features. Panther-Bridge applies effective visual token pruning, reducing redundant information and lowering computational costs in multi-turn training. Finally, the Panther-Decoder is trained in interleaved mode and compatible with a variety of LLMs. 
Experimental results show that our method consistently outperforms existing approaches on various benchmarks, particularly in vision-centric tasks. This work not only uncovers a significant issue in current MLLMs but also provides an effective solution, laying the groundwork for future research and applications.

{
    \small
    \bibliographystyle{ieeenat_fullname}
    \bibliography{main}
}

\input{sec/X_suppl}

\end{document}

%% file: sec/0_abstract.tex
\begin{abstract}

Multimodal large language models (MLLMs) are closing the gap to human visual perception capability rapidly, while, still lag behind on attending to subtle images details or locating small objects precisely, etc. Common schemes to tackle these issues include deploying multiple vision encoders or operating on original high-resolution images. Few studies have concentrated on taking the textual instruction into improving visual representation, resulting in losing focus in some vision-centric tasks, a phenomenon we herein termed as Amblyopia. In this work, we introduce Panther, a MLLM that closely adheres to user instruction and locates targets of interests precisely, with the finesse of a black panther. Specifically, Panther comprises three integral components: Panther-VE, Panther-Bridge, and Panther-Decoder. Panther-VE integrates user instruction information at the early stages of the vision encoder, thereby extracting the most relevant and useful visual representations. The Panther-Bridge module, equipped with powerful filtering capabilities, significantly reduces redundant visual information, leading to a substantial savings in training costs. The Panther-Decoder is versatile and can be employed with any decoder-only architecture of LLMs without discrimination. Experimental results, particularly on vision-centric benchmarks, have demonstrated the effectiveness of Panther.

\end{abstract}

%% file: sec/1_intro.tex
\section{Introduction}
\label{sec:intro}
\begin{figure}[htbp]
\centering
\includegraphics[width=1.\linewidth]{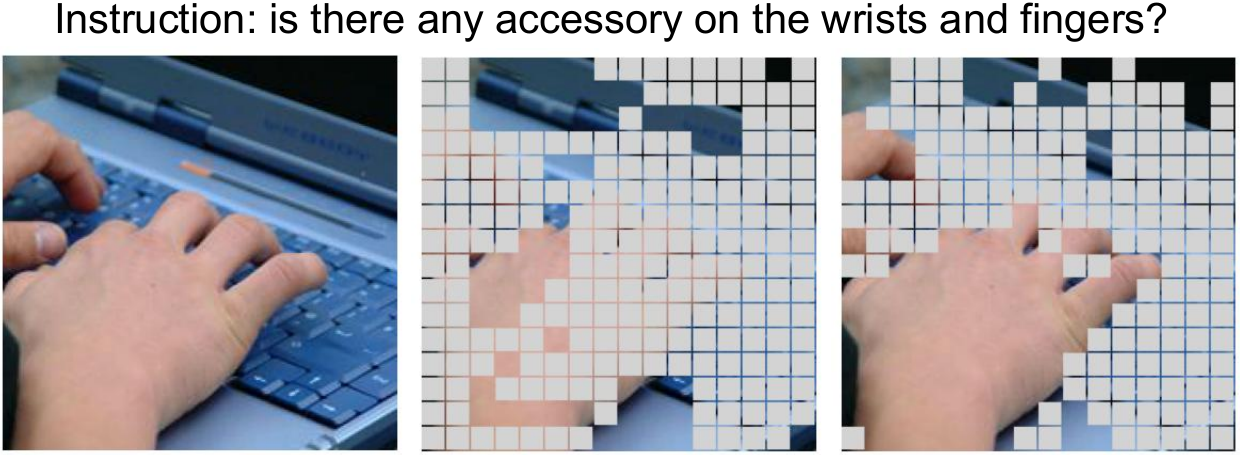} \\
\caption{A comparative analysis of visual feature heatmaps between LLaVA (center) and our advanced Panther (right) with deep text-instructed visual prompting. Panther distinctly enhances focused attention on the visual elements targeted by the instruction.}
\label{fig1}
\vspace{-4mm}
\end{figure}
Multimodal large language models (MLLMs) have gained significant attention in the research community, serving as foundational elements for the development of general-purpose assistants~\citep{gpt4v,alayrac2022flamingo,zhu2023minigpt}. Most studies on MLLMs leverage pretrained vision encoders~\citep{clip,zhai2023sigmoid} and focus on enhancing vision-language connectors~\citep{llava,alayrac2022flamingo,li2023blip} and scaling up the LLM decoders~\citep{lu2023empirical,touvron2023llama,chatgptplugins}.Training datasets, particularly those optimized for visual instruction tuning \citep{llava}, have further bolstered the models' proficiency in adhering to natural instructions \citep{liu2023mmbench,yu2023mmvet,li2023seed,fu2023mme} and in executing visual reasoning tasks \citep{hudson2019gqa,goyal2017vqav2,gurari2018vizwiz}, as reflected in their benchmark performances.

However, recent vision-centric benchmarks, such as MMVP and Realworld-QA~\citep{tong2024mmvp,tong2024cvbench,xai2024grokv}, have highlighted systematic visual impairments in the capabilities of MLLMs, particularly in tasks involving spatial relationships and object counting. In response, several studies~\citep{liu2024improved,li2024monkey,dong2024internlm,tong2024cvbench} have focused on enhancing the visual modules of MLLMs to address these limitations. These approaches include increasing input image resolution~\citep{liu2024improved,dong2024internlm} to capture finer visual details~\citep{li2024monkey} and integrating multiple visual encoders~\citep{lin2023sphinx,tong2024cvbench,lu2024deepseek}, as each encoder type~\citep{clip,zhai2023sigmoid,oquab2023dinov2,kirillov2023segment} emphasizes distinct visual features~\citep{tong2024mmvp}.

\begin{figure*}[htbp]
\centering
\includegraphics[width=1.\linewidth]
{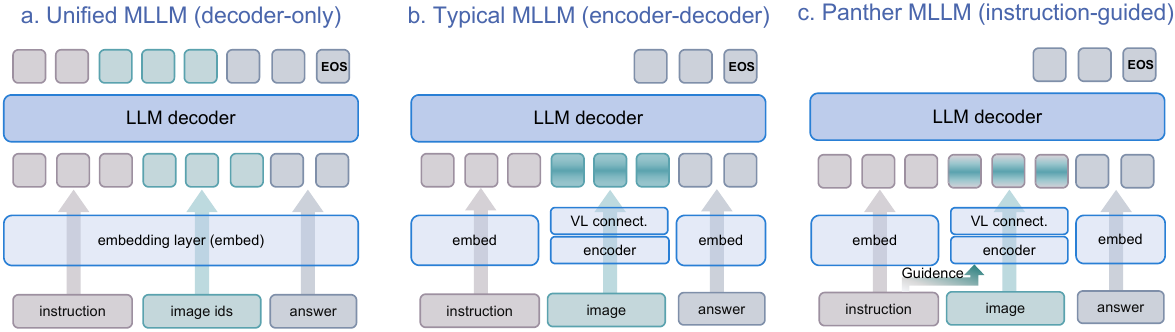} \\
\caption{Comparison of MLLM Architectures: (a) \textbf{Unified MLLM (decoder-only)} enables early fusion of instructions and images, preserving low-level image features before fusion but lacks pretrained vision-language knowledge.
(b) \textbf{Typical MLLM (encoder-decoder)} performs late fusion, leading to the \textit{Amblyopia} issue where important visual details may be filtered out.
(c) \textbf{Our Panther MLLM} convert instructions as visual prompts to guide the visual encoder for extracting instruction-aware feature. Panther strikes a balance between enhancing instruction-specific visual features and retaining the knowledge acquired from pre-training.}
\label{fig2}
\vspace{-3mm}
\end{figure*}

Despite these efforts, a persistent issue remains unresolved: certain visual details sought by users may be overlooked due to the absence of integrated instructions and the excessive number of vision block layers. This situation consequently leads to an overly diffused model focus, as illustrated in Figure  \ref{fig1}, where the spatial features on the fingers are not clearly captured, and the wrist features appear dispersed. We term this phenomenon \textit{Amblyopia}. In essence, this issue originates from the lated integration of image and text instructions within the architecture of encoder-decoder MLLMs, as depicted in Figure  \ref{fig2}b. Although early fusion in decoder-only MLLMs \cite{gemini2024,fuyu8b2024,zhan2024anygpt,Chameleon2024earlyfusion,xie2024showo,wu2024vila,wang2024miofoundationmodelmultimodal}, as shown in Figure  \ref{fig2}a, is more feasible but they still lags behind encoder-decoder MLLMs in visual comprehension. This gap is attributed to the lack of pre-trained vision-language knowledge \cite{radford2021clip}, as evidenced by \cite{xulibra} and \cite{wu2024vila}.  These models \cite{xulibra,wu2024vila} typically require about ten times the training data of LLaVA \cite{llava} to achieve comparable performance. Therefore, we adhere to employing the encoder-decoder architecture.

In this paper, we present Panther, an advanced MLLM that demonstrates exceptional proficiency in following user instructions and accurately pinpointing objects within image, akin to the purposefulness and precision of a black panther. Panther is constructed with three core modules: Panther-VE (visual encoder), Panther-Bridge and Panther-Decoder. Specifically, within Panther-VE, we utilize a lightweight encoder to transform user instructions into learnable prompts, which are subsequently integrated into the vision encoder to yield adaptive, instruction-guided visual features, as illustrated in Figure  \ref{fig2}c. Although several works \cite{cai2024vipllava,lin2024drawandunderstand,jiao2024enhancing,lin2024vp_seg} also attempt to incorporate visual prompts in MLLMs, they overlook the significance of textual instructions. The Panther-Bridge module is meticulously designed to handle multi-turn queries by filtering out redundant visual tokens across different rounds, ensuring that only the most relevant and precise information is retained. This approach substantially mitigates the computational expense during the training stage. For Panther-Decoder,  we have tailored the training scheme to an interleaved mode to accommodate Panther-VE, and Panther exhibits versatility in its compatibility with any decoder-only architectures of LLMs.

Experiments spanning a diverse benchmarks have validated the effectiveness of Panther, encompassing visual question-answering, instruction-following, and vision-centric tasks. Our main contribution are three folds:

\begin{enumerate}
    \item  We introduce Panther, a novel framework designed to mitigate the \textit{Amblyopia} phenomenon prevalent in encoder-decoder MLLMs. 
    \item  Panther-VE has been introduced to integrate user instructions with the visual representation of images, thereby generating instruction-aware visual embeddings. 
    \item The Panther-Bridge serves as an intermediary between the Panther-VE and Panther-LLM, significantly filtering out redundant tokens in multi-turn scenarios, thereby substantially reducing the training cost overhead.
\end{enumerate}

%% file: sec/2_related_work.tex
\section{Related Work}
\subsection{Multimodal Large Language Models (MLLMs)}
Typical architectures \cite{llava,li2023blip,zhu2023minigpt,dai2023instructblip,bai2023qwen} of MLLMs consist of three main components: a pre-trained visual backbone for encoding visual features, a pre-trained LLM for understanding user instructions and generating responses, and a vision-language cross-modal connector to bridge the output of the visual encoder with the language model. 
The training of MLLMs \cite{llava,zhu2023minigpt} usually can be divided into two stages. The first stage is vision-language pre-training, which aligns visual features with the language model’s word embedding space using image-text pairs. 
The second stage, visual instruction tuning, fine-tunes the model on visual instructions~\citep{llava} to handle a wide range of tasks that involve visual content. 
Recently, unified MLLMs (decoder-only) \cite{liu2024worldmodelmillionlengthvideo,xie2024showo,wu2024vila,Chameleon2024earlyfusion,ge2024seed} take both visual and text tokens as unified discrete entity to attain visual understanding and generation simultaneously. However, their performance in visual comprehension lags behind the typical MLLMs due to the abandonment of pretrained vision-language knowledge~\citep{radford2021clip,xulibra,wu2024vila}. Thus, these MLLMs need more data for pretraining to close this gap. 
In this paper, we mainly focus on the typical MLLMs with visual-encoder and LLM-decoder for their training efficiency and visual understanding performance.

\subsection{Visual Impairment of MLLMs}
Despite demonstrating promising performance on standard QA benchmarks~\citep{hudson2019gqa,goyal2017vqav2,gurari2018vizwiz} and instruction-following tasks~\citep{yu2023mmvet,liu2023mmbench,fu2023mme}, recent studies~\citep{wang2023gvt,tong2024mmvp,xai2024grokv,tong2024cvbench} highlight the visual impairment exists in MLLMs. 
GVT~\citep{wang2023gvt} and MMVP~\citep{tong2024mmvp} both reveal distinct strengths and limitations of CLIP \cite{radford2021clip,zhai2023sigmoid} and DINO \cite{oquab2023dinov2} within MLLMs. GVT contributes by establishing a benchmark for low-level perception, while MMVP pinpoints specific instances where CLIP encounters failures. To further explore these limitations, CV-Bench~\citep{tong2024cvbench} and Realworld-QA~\citep{xai2024grokv} extend existing benchmarks to a larger scale. In order to mitigate the visual impairment, recent studies~\citep{wang2023gvt,tong2024cvbench,lu2024deepseek} incorporate multiple vision encoders such as CLIP, DINO and SAM~\citep{kirillov2023segment}. Additionally, some works~\citep{liu2024improved,dong2024internlm,li2024monkey,wang2024qwen2vl,yao2024minicpmv} adopted higher or adaptive resolution image inputs to enhance visual acuity. However, these methods, which rely on fixed visual encoders, tend to become dispersed and overlook the necessity to maintain focus on the instructions, a phenomenon we termed \textit{Amblyopia}. Q-former~\citep{li2023blip} attempts to cross-attend to visual features based on instructions, but it only operates on the final feature layer.
In this paper, we tackle the \textit{Amblyopia} issue of MLLMs by applying instruction guidance to prompting the visual encoder from shallow layers.

\subsection{Prompt Learning} 
Originally developed for Transformer-based LMs~\citep{vaswani2017attention,devlin2018bert}, prompt learning has grown from manual text prompts for in-context learning \citep{radford2019language,brown2020language} to automated and gradient-based soft prompts for parameter-efficient tuning \citep{shin2020autoprompt,prompt-tuning,prefix-tuning,p-tuning}. Inspired by NLP, this technique has been applied to Vision Transformers \citep{vit,jia2022visual} downstream domains ~\citep{ge2023domain,wang2022l2p,chen2023prompt,Li_2023_CVPR,li2024longmil,10377922,shui2024prompt}. For multimodal vision-language models, soft prompts enhance few-shot learning, though mostly with randomly initialized text prompts~\citep{radford2021clip,coop,khattak2023maple}. In the context of MLLM, \citep{cai2024vipllava} propose using circle marks as visual prompts to facilitate the integration of human intentions \citep{shtedritski2023does}. \citep{lin2024drawandunderstand} introduce extra data for learning point and bounding box prompts for pixel-level and region-specific understanding. \citep{jiao2024enhancing,lin2024vp_seg} integrate fine-grained knowledge from instance segmentation models as visual prompts. However, these methods overlook user intention in textual instructions. Unlike prior work, we generate visual prompts by projecting text instruction embeddings into the visual space, enabling the visual encoder to focus on user-specified entities in the instructions, thereby enhancing focus on critical visual details.

%% file: sec/3_method.tex
\section{Method}
This section outlines our paper structure. Section \ref{MLLM} reviews MLLM and Prompt-Tuning definitions. Section \ref{framework} introduces the Panther framework, with Panther-VE in Section \ref{tpi} detailing instruction-aware visual prompts. Section \ref{token_prune} covers Panther-Bridge, improving multi-turn QA training efficiency, while Section \ref{decoder} describes training instruction-aware visual features in the Panther-Decoder. Finally, the inference details are provided in Section \ref{sec:inference}.

\subsection{Preliminaries} \label{MLLM}
\paragraph{MLLMs:}
The aim of MLLMs is to create models capable of generating responses from multimodal inputs, including both visual and textual data. The typically encoder-decoder MLLMs consist of three main components:

\begin{itemize}
\item Visual Encoder $\mathbf{F}_{I}$ converts an input image ${I}_{img} \in \mathbb{R}^{H \times W \times 3}$ into visual embeddings $ {I}_v \in \mathbb{R}^{N \times d}$. For instance, using CLIP-ViT-L/14 as the backbone with patch size $P=14$, $N = HW / P^2$, and $d$ is the dimension.
\item Visual-Language Connector $\mathbf{\Gamma}_{I \rightarrow T}$ maps the visual embeddings ${I}_v$ into visual tokens $\mathbf{T}_v$ within the textual embedding space, aligning dimensions ($d \rightarrow d_1$) for the language model's input.
\item LLM Decoder ${\mathbf{\Phi}({\mathbf{T}_v},{\mathbf{T}_t})}$ takes both visual tokens $\mathbf{T}_v$ and textual tokens $\mathbf{T}_t$, and auto-regressively generates a coherent response. For a sequence of responses with length $M$, the probability of generating contextually relevant answers $\mathbf{Y}=\{y_i\}^M_{i=1}$ is calculated as follows:
\end{itemize}
\begin{equation}
\label{eq:llm_ar}
p(\mathbf{Y}|{\mathbf{T}_{v}},{\mathbf{T}_{t}}) = \prod_{i=1}^{M}p(y_{i}|{\mathbf{T}_{v}},{\mathbf{T}_{t}^{<i}}, {\mathbf{Y}^{<i}} ) .
\end{equation}
\paragraph{Visual Prompt Tuning:}  
Considering the $j$-th Transformer layer ($L_{j}$) of ViT with input image patch embedding tokens (${I}_{j-1} \in \mathbb{R}^{N \times d} $), together with an extra class token \texttt{[CLS]} ($C_{j-1}  \in \mathbb{R}^{1 \times d} $), the forward of a Transformer layer can be formulated as:
\begin{equation}
\label{eq:vit}
[C_{j}, {I}_j] = L_j( [C_{j-1}, {I}_{j-1}]),
\end{equation}
where $[\cdot,\cdot]$ indicates concatenation on the sequence dimension, thus $[{C}_{j}, {I}_{j}]\in\mathbb{R}^{(1+N)\times d}$. 
\begin{figure*}[htbp]
\centering
\includegraphics[width=.9\linewidth, ]{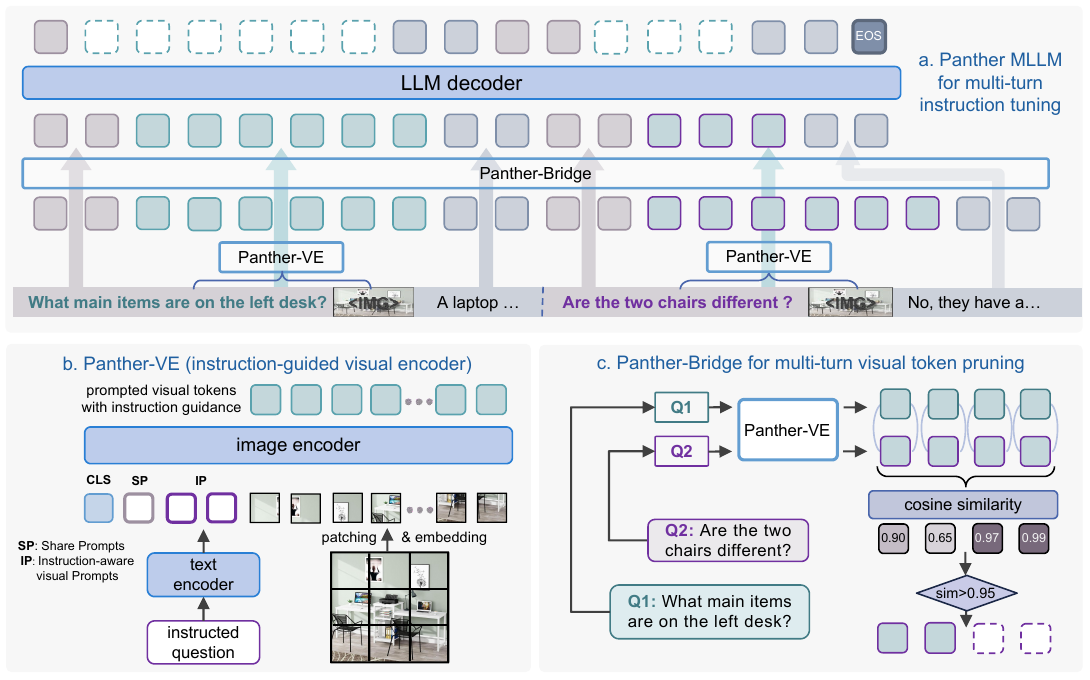} \\
\caption{ The overall framework. (a) The Panther MLLM instruction tuning on multi-turn visual QA data, the visual tokens are generated via Panther-VE and pruned via Panther-Bridge. (b) The Panther-VE, aiming at introducing instruction guidance on visual feature, generates instruction-aware visual prompts to focus specific image details. (c) The Panther-Bridge for multi-turn visual tokens pruning, here we only show two turn case as example. }
\label{fig3_framework}
\vspace{-4mm}
\end{figure*}
In the context of prompt tuning on a pretrained ViT, a set of \texttt{[PROMPTS]} ($P_{j-1}  \in \mathbb{R}^{K \times d} $) are pre-pended to ${I}_{j-1}$ the sequence dimension in each layer $j$. 
Then, during fine-tuning, the forward pass of a Transformer layer can be formulated as:
\begin{align}
\label{eq:shallow}
    [{C}_{j}, \cancel{P}_{j}, {I}_{j}] = {L_i}([{C}_{j-1}, {P}_{j-1},{I}_{j-1}]) .
\end{align}
Note that here we only formulate the deep visual prompts scheme \cite{vpt}, where the prompts in each layer are random-initialized learnable vectors and the correspondence output tokens $\cancel{P}_{j}$ will be discarded. The learnable prompts are updated during fine-tuning, while the Transformer backbone remains frozen. 

\subsection{Panther} \label{framework}

As shown in Figure  \ref{fig3_framework}, the overall panther framework contains three key components: Panther-VE, Panther-Bridge and Panther-Decoder. For a multi-turn text-and-image conversation $\{ {I}_{img}, q_0, a_0, q_1, a_1, ... , q_k, a_k , ... , q_{K-1}, a_{K-1}\}$ ($K$ turns in total and $k \in [0, K-1]$), the initial step involves the utilization of Panther-VE to extract instruction-aware image representations from each individual turn of the conversation. Subsequently, the information gleaned from these multi-turn interactions is subjected to Panther-Bridge, which serves to filter out redundant and superfluous elements, thereby reducing the computational load. Finally, the condensed information from these multi-turn dialogues is fed to Panther-Decoder to facilitate the prediction of subsequent tokens. In the subsequent sections, we will provide a comprehensive description of the specific methodologies employed within each module.

\subsubsection{Panther-VE} \label{tpi}
Here, we provide a detailed exposition on the specific implementation of Panther-VE. Through the integration of user instruction information, Panther-VE generates visual representations that are more finely tuned to the regions of interest specified by the user. Like panther, it possesses a strong aptitude for auditory (instruction-following) and visual perception capabilities.

Specifically, for each $\{ {I}_{img}, q_k, a_k\}$, within the Panther-VE framework, we initially employ a lightweight text-encoder, denoted as $\mathbf{F}_{T} $ (for instance, BERT \citep{devlin2018bert}), to transform the textual instructions into a text embedding. Subsequently, a straightforward multilayer perception, represented by $\mathbf{\Gamma}_{T\rightarrow I}$ is utilized to project the text embedding into the visual space. Then the instruction-aware visual prompts is pre-pended to the visual embedding. The generation of instruction-aware visual prompts can be formatted as $\mathbf{\Gamma}_{T\rightarrow I} [ \mathbf{F}_{T}(q_k)]$.

Additionally, since the vision encoder has not previously encountered instruction-aware visual prompts, direct training can be relatively challenging. Therefore, to alleviate this training complexity, we also introduce several randomly initialized visual prompts to address this issue. We designate these prompts as \textit{shared prompts (sp)} due to their application across all samples, which standardizes the training process. Thus, for each $\{ {I}_{img}, q_k, a_k\}$,  the process of extracting visual tokens through Panther-VE can be represented as follows:
\begin{equation}
     \mathbf{T}_{v}^{\prime}= \ \mathbf{\Gamma}_{I\rightarrow T} \{ \mathbf{F}_{I}({I}_{img}, \ sp, \mathbf{\Gamma}_{T\rightarrow I} [ \mathbf{F}_{T}(q_k)   ] ) \} .
\label{eq:panther_ve}
\end{equation}


\subsubsection{Panther-Bridge} \label{token_prune}
In a typical MLLM framework, the computation and memory demands are primarily driven by the LLM ${\mathbf{\Phi}_{({\mathbf{T}_v},{\mathbf{T}_t})}}$, which contains a large number of parameters. Notably, the computational cost of LLMs increases quadratically with the number of input tokens.
In this subsection we first elaborate on how LLaVA processes multi-turn visual QA training data during instruction-tuning. Then, we explain why our Panther-VE would cause the long sequence problem and how to use Panther-Bridge to solve it.
\paragraph{LLaVA for multi-turn:}
The visual features for the image are only need to be computed once in LLaVA given multi-turn data $\{ {I}_{img}, q_0, a_0, ... , q_k, a_k , ... , q_{K-1}, a_{K-1}\}$ since:
\begin{equation}
    {I_v} = I_v^{k} = I_v^{k+1} = \mathbf{F}_{I}({I}_{img}),
\end{equation}
then, ${I_v}$ is converted into textual embedding space (size $d_1$) as  $\mathbf{T}_v \in \mathbb{R}^{N \times d_1}$, and the remained texts of $q_k,a_k$ are tokenized and embedded into $\mathbf{T}^e_t = [ T_t^{q_0}, T_t^{a_0}, ..., T_t^{q_{K-1}}, T_t^{a_{K-1}} ] \in \mathbb{R}^{M^{\prime} \times d_1}$ ($M^{\prime}$ is the overall textual tokens length). 
The overall tokens' embedding sequence input into LLM decoder is:
\begin{equation}
 [\mathbf{T}_v, \mathbf{T}^e_t] \in \mathbb{R}^{(M^{\prime}+N) \times d_1},
\end{equation}
note that here we omit the system prompts for simplicity and the overall token length $M^{\prime}+N \leq 2048$ is true for almost all data given $N=576$.   

\paragraph{Long sequence dilemma:}
Our Panther-VE generates unique visual embeddings for each text instruction in multi-turn QA data. This approach, while enhancing relevance to specific instructions, may lead to excessively long training sequences within the context of multi-turn interactions.
The prompted visual embeddings can be generated via Panther-VE in Eq. \ref{eq:panther_ve} as:
\begin{align}
    {I}^{k} &= \mathbf{F}_{I}({I}_{img}, \ sp, \ \mathbf{\Gamma}_{T\rightarrow I} [ \mathbf{F}_{T}(q_k)   ] ), \\  \nonumber
    {I}^{k+1} &= \mathbf{F}_{I}({I}_{img}, \ sp, \ \mathbf{\Gamma}_{T\rightarrow I} [ \mathbf{F}_{T}(q_{k+1})   ] ),
\end{align}
where $k$ and $k+1$ represent the order of dialogue turns, each associated with distinct question $q_k$ and $q_{k+1}$. When these questions are applied to the same image, they elicit two distinct image features, ${I}_{}^{k}$ and ${I}_{}^{k+1}$.
After being projected into the textual embedding space, all visual tokens for the LLM can be represented as follows: $[\mathbf{T}_v^{\prime_0}, \mathbf{T}_v^{\prime_1}, ... , \mathbf{T}_v^{\prime_{K-1}} ] \in \mathbb{R}^{NK \times d_1}$, and the sequence of embedding tokens fed into the LLM decoder is: 
\begin{align}
    [\mathbf{T}_v^{\prime_0}, T_t^{q_0}, T_t^{a_0},  \mathbf{T}_v^{\prime_{1}}, 
 T_t^{q_1}, T_t^{a_1}, 
 ... , &\mathbf{T}_v^{\prime_{K-1}}, T_t^{q_{K-1}}, T_t^{a_{K-1}} ] \nonumber \\ 
 &\in \mathbb{R}^{(M^{\prime}+NK) \times d_1}.
\end{align}
Obviously, it is highly probable that the total token length $M^{\prime}+NK$ will exceed 2048. For instance, when the total number of turns $K \ge 4$ with $N=576$, the inequality $M^{\prime}+NK>NK \ge 576*4 = 2304 > 2048$ holds true. This emphasizes the necessity of enhancing training efficiency within our framework.

\paragraph{Panther-Bridge for token pruning:}
To address the challenge of overly long training sequences in multi-turn data, we introduce Panther-Bridge for token pruning. 
Without loss of generality, we consider a two-turn QA setting.
Panther-Bridge aims to prune $\mathbf{T}_v^{\prime_1}$ in sequence by the metric of cosine similarity, which can be calculated via $ \cos(T_{v,i}^{\prime_0}, T_{v,i}^{\prime_1}) = \frac{T_{v,i}^{\prime_0} \cdot T_{v,i}^{\prime_1}}{\|T_{v,i}^{\prime_0}\| \|T_{v,i}^{\prime_1}\|} $, where $T_{v,i}^{\prime_0} \in \mathbb{R}^{1 \times d_1} $ is the $i$-th element of $\mathbf{T}_{v}^{\prime_0}\in \mathbb{R}^{N \times d_1} $ along the sequence dimension.
Then, we prune the tokens in $T_{v}^{\prime_1}$ if the cosine similarity exceeds a threshold $\tau$ and retain the useful tokens as:
\begin{equation}
\mathbf{T}_{v,\text{useful}}^{\prime_1} = \{ T_{v,i}^{\prime_1} \mid \cos(T_{v,i}^{\prime_0}, T_{v,i}^{\prime_1}) \leq \tau \}.
\label{eq:cos_prune}
\end{equation}
Similarly, when the maximum number of turns is three, we can further prune $ \mathbf{T}_{v}^{\prime_2} $ based on its similarity to $ \mathbf{T}_{v}^{\prime_0} $, as shown in Eq. \ref{eq:cos_prune}, resulting in $ \mathbf{T}_{v,\text{useful}}^{\prime_2} $. If there are common spatial token indices (with the original index ranging from $ 0 $ to $ N-1 $) between $ \mathbf{T}_{v,\text{useful}}^{\prime_2} $ and $ \mathbf{T}_{v,\text{useful}}^{\prime_1} $, the pruning process will further be applied to their corresponding tokens. This process can be generalized to $K$ rounds conversation, with the detailed procedures outlined in the Appendix.

We forgo the token length analysis in this context, as the relationship between the pruning ratio and the threshold $\tau$ is not amenable to direct computation. Instead, we concentrate on an empirical examination of the training time expenditure. Please refer to Table \ref{tab:token_prune} for detailed results.

\subsubsection{Panther-Decoder} \label{decoder}
To facilitate multi-turn training with our distinct visual features of each turn (as illustrated in Section \ref{token_prune}), the Panther's LLM decoder are trained via an interleaved mode. The first turn ($k=0$) is just as same to the original mode in Eq. \ref{eq:llm_ar}:
\vspace{-5mm}
\begin{equation}
     p(\mathbf{Y}|{\mathbf{T}_{v}^{\prime_k}},{\mathbf{T}_{t}}) = \prod_{i=1}^{M_0}p(y_{i}|{\mathbf{T}_{v}^{\prime_k}},{\mathbf{T}_{t}^{<i}}, {\mathbf{Y}^{<i}} )  ,
\vspace{-2mm}
\end{equation}
where the $M_0$ is the sequence length of QA in first turn. Similarly, the auto-regressive training in second turn after token pruning in Section  \ref{token_prune} can be derived as:  
\begin{equation}
\begin{aligned}
    & p(\mathbf{Y}|\mathbf{T}_{v}^{\prime_k}, \mathbf{T}_{v,useful}^{\prime_{k+1}}, \mathbf{T}_t) \\ = & \prod_{i=M_0+1}^{M_1} p(y_i | \mathbf{T}_{v}^{\prime_k}, \mathbf{T}_{v,useful}^{\prime_{k+1}}, \mathbf{T}_t^{<i}, \mathbf{Y}^{<i}),
    \end{aligned}
\end{equation}
where the textual tokens of instruction $q_{0}$ and answer $a_{0}$ are included in  $\mathbf{T}_t^{<i}$ considering initial case $i=M_0+1$. Moreover, our approaches including Panther-VE, Panther-Bridge and interleaved training scheme are orthogonal to the implementation mechanisms of LLMs, thus could support various kinds of LLMs as validated in experiments Section  \ref{sec_ablation} and Table \ref{tab:LLM}.

\subsection{Inference} \label{sec:inference}
During inference, Panther-VE and Panther-Decoder operate in the same manner as the training setting described in Section  \ref{framework}. However, for optimal evaluation performance, we omit the Panther-Bridge module. In other words, we does not make pruning but input the original visual token embedding $T_{v}^{\prime_{k+1}}$ in Section  \ref{token_prune} to LLM.

%% file: sec/4_experiments.tex
\section{Experiments}

\begin{table*}[h]
\centering
\scalebox{0.85}{
\begin{tabular}{l l c c c | c c c c | c c c }
\toprule
\multirow{2}{*}{\textbf{Method}} & \multirow{2}{*}{\textbf{LLM}} & \textbf{Image} & \multicolumn{2}{c|}{\textbf{Sample Size}} & \multirow{2}{*}{\textbf{VQA$^{\textnormal{v2}}$}} & \multirow{2}{*}{\textbf{GQA}}  & \multirow{2}{*}{\textbf{SQA$^\textnormal{I}$}} & \multirow{2}{*}{\textbf{VQA$^\textnormal{T}$}} & \multirow{2}{*}{\textbf{POPE}} & \multirow{2}{*}{\textbf{MME}} & \multirow{2}{*}{\textbf{MMB}}\\
 & & \textbf{Size} & \textbf{Pretrain} & \textbf{Finetune} &  &  &  &    &   \\
\midrule
BLIP-2 & Vicuna-13B & 224$^2$ & 129M & - & 65.0 & 41.0  & 61.0 & 42.5 &  85.3 & 1293.8 & --\\
InstructBLIP & Vicuna-7B & 224$^2$ & 129M & 1.2M & --  & 34.5 & 60.5 & 50.1 & --  & -- & 36.0\\
InstructBLIP & Vicuna-13B & 224$^2$ & 129M & 1.2M & --  & 33.4 & 63.1 & 50.7 &  78.9 & 1212.8 & --\\
Qwen-VL & Qwen-7B & 448$^2$ & 1.4B & 50M & 78.8 & 59.3  & 67.1 & \underline{63.8} & -- & -- & 38.2 \\
Qwen-VL-Chat & Qwen-7B & 448$^2$ & 1.4B & 50M & 78.2& 57.5  & 68.2 & 61.5 & -- & 1487.5 & 60.6\\
mPLUG-Owl2 & Llama-7B & 448$^2$ & 400M & 1.2M & 79.4 &56.1   & - & 58.2 & 85.8 & 1450.2 & 64.5 \\
VILA & Llama 2-7B & 336$^2$ & 50M & 1M & 79.9 & 62.3  & 68.2 & \textbf{64.4} & 85.5 & \textbf{1533.0} & 68.9 \\
\midrule
LWM &  Llama-2-7B & 256$^2$ & - & - & 55.8 & 44.8 & -& 18.8 & 75.2 & - & - \\
Show-o & Phi-1.5-1.3B & 256$^2$ & - & - & 59.3 & 48.7 & -& - & 73.8 & 948.4 & - \\
VILA-U & Llama-2-7B & 384$^2$ & - & 7M & 79.4 & 60.8 & -& 60.8 & 85.8 &1401.8 & - \\

\midrule

LLaVA-TokenPack & Vicuna-7B &  336$^2$&  558K & 665K  &  77.9 &  61.9  & -  & 57.2 & \underline{87.0} & -& 65.1 \\
Cambrian-dev & Vicuna-7B & $^{\dagger}$518$^2$ & 1.2M & 737K &  - & 63.3  & 68.8 & 60.4 & - & 1432.0 & 61.3 \\
LLaVA-VisPrompt  &  Vicuna-7B &  336$^2$&  558K & 665K  &  79.8 &  63.3  & 69.5  & 59.8 & \textbf{88.9} & \underline{1515.3} & 67.6 \\

\newshortname{} & Vicuna-7B & 336$^2$ & 558K & 665K & 78.5 & 62.0  & 66.8 & 58.2 & 85.9 & 1510.7 & 64.3 \\
\rowcolor{cyan!20}
 \ourshortname{} & Vicuna-7B & 336$^2$ & 558K & 665K &  \underline{80.8} & 	\underline{65.2} &	67.8&	59.6 & 86.7 & 1507.3 & 67.1\\
 \newshortname{} &  Llama 3-8B & 336$^2$ & 558K & 665K & 79.7 & 63.3  & \underline{73.3} & 58.4 & 84.6 & 1506.5 &  \underline{68.9} \\
 
\rowcolor{cyan!20}
 \ourshortname{} & Llama 3-8B & 336$^2$ & 558K & 665K & \textbf{81.2} & 	\textbf{65.5} &	\textbf{73.5} &	59.5   &  85.7 & 1513.6  & \textbf{71.9}\\
 
\bottomrule
\end{tabular}
}
\caption{Comparison with SoTA methods on \textbf{general VQA} and \textbf{instruction-following benchmarks}. \textbf{Bold} represents the highest, and \underline{underlined} represents the second highest. The first row shows previous popular MLLMs, the second row includes decoder-only MLLMs, the third row includes our method and comparisons to MLLMs with similar structures. Note that Cambrian-dev combines multiple visual encoders, the largest one is $\dagger$DINOv2-ViT14@518. 
}
\vspace{-3mm}
\label{tab:results1}
\end{table*}

In this section, we conduct comprehensive comparisons of our method with existing state-of-the-art multimodal models. Additionally, we perform a series of ablation studies to further validate the proposed method. Finally, we provide visualization examples for in-depth analysis.

\subsection{Models and Training Settings}
We employ the same setting as LLaVA-1.5~\citep{llava} with CLIP-ViT-L/p14-336px~\citep{clip} as the default vision encoder, Vicuna-7B~\citep{vicuna} model as the default language decoder and a two-layer MLP as the multimodal projector. The text encoder from CLIP is leveraged to extract the textual embeddings of instructions. We adopt the same pre-training alignment and instruction-tuning settings as LLaVA-1.5~\citep{llava}, utilizing the same datasets (558K samples for pre-training and 665K for instruction-tuning), batch size, and learning rate. 

Panther is activated only during the instruction-tuning stage, where Panther-VE exclusively updates the shared prompts (24 prompts as default) and the instruction-aware visual Prompt's (IP) generator  using a base learning rate of 1e-4, while keeping the original ViT parameters frozen. The generator of IP consist of the MLP projector (a two-layer linear module, updated) and the CLIP's text encoder (frozen). Note that we generate 77 IP with mask based on the max-length of CLIP's text encoder and instructions practical length. In addition, the default value of $\tau$ is 0.95 in Panther-Bridge. Similar to LLaVA-1.5, the vision-language connector and LLM decoder are also updated. All the models are trained on 8 $\times$ NVIDIA A100 GPUs.

\subsection{Eval Benchmarks and Compared Baselines}
We evaluate MLLMs on 10 benchmarks including general VQA (VQA-v2, GQA, ScienceQA and TextVQA~\citep{goyal2017vqav2,hudson2019gqa,lu2022learn,singh2019textvqa}), instruction-following benchmarks (POPE, MME-Perception and MMBench~\citep{li2023pope,fu2023mme,liu2023mmbench}) and vision-centric benchmarks:
MMVP~\citep{tong2024mmvp} evaluate on the curated VQA image that CLIP worse than DINO. CV-Bench~\citep{tong2024cvbench} focus on the 2D spatial relationship and object counting, also on 3D depth and distance. Realworld-QA (RWQA)~\citep{xai2024grokv} evaluate the spatial understanding capabilities.

%
We compare our results with a bunch of state-of-the-art open-source MLLMs with similar training data size and input resolution:\\
1) General MLLMs:
BLIP-2~\citep{li2023blip}, InstructBLIP~\citep{dai2023instructblip}, Qwen-VL~\citep{bai2023qwen}, mPLUG-Owl2~\citep{ye2024mplug}, LLaVA-1.5~\citep{llava}, VILA~\citep{lin2023vila}. \\
2) Visual-centric MLLMs: Cambrian~\citep{tong2024cvbench} combines multiple visual encoders and LLaVA-TokenPacker~\citep{li2024tokenpacker} combines multiple layers input of visual encoder. \\
3) LLaVA-1.5 with visual prompt: LLaVA-VisPrompt~\citep{lin2024vp_seg}. 
4) Decoder-only MLLMs including LWM~\citep{liu2024worldmodelmillionlengthvideo}, Show-o~\citep{xie2024showo} and VILA-U~\citep{wu2024vila} with similar LLM size to ours.

\subsection{Main Results}

\paragraph{General QA and Instruction-Following Benchmarks:}

The results are shown in Table \ref{tab:results1}, it can be seen that our \ourshortname{} surpasses previous models in 4 out of 7 benchmarks and showing consistency improvement compared to the baseline (\newshortname{}). Specifically, \ourshortname{} achieves state-of-the-art performance on the VQAv2 and GQA, where even \underline{\ourshortname{} with Vicuna-7B}  can surpass \underline{\newshortname{} with Llama 3-8B \cite{dubey2024llama3herdmodels}}. This reveals that even more powerful LLMs cannot address all types of evaluation samples effectively, there remains a necessity for high-quality visual representations.
Averagely, our Panther improves the performance  approximately 1.92\% and 1.52\% for Vicuna-7B and Llama 3-8B respectively compared to \newshortname{} (MME is excluded for its different value scale). These results highlight the effectiveness of Panther, indicating its robustness and adaptability in handling complex visual-question tasks.

\paragraph{Visual-Centric Benchmarks:} 
To better validate Panther's visual perception capability, we evaluate it on visual-centric benchmarks and the results are shown in Table \ref{tab:results2}. Since most pervious models do not evaluate the performance of these benchmarks, we mainly compare our method with Cambrian~\citep{tong2024cvbench}. Note that Cambrian-1 combines 4 visual encoders and are trained on 2.5M (pretrain) + 7M (fine-tuning) data. The Cambrian-dev (from their ablations) combines only 2 visual encoders and trained with similar data as ours, thus can be treated as a fair baseline. Our \ourshortname{} surpasses Cambrian-dev in all the 4 benchmarks, and shows consistency improvement compared to \newshortname{}. 
Averagely, the performance of Panther improves approximately 2.90\% and 3.15\% with Vicuna-7B and Llama 3-8B as LLM respectively when compared to \newshortname{}.

\begin{table}[htbp]
\centering
\scalebox{0.85}{
\begin{tabular}{l c | c c  c c   }
\toprule
\multirow{2}{*}{\textbf{Method}} & \multirow{2}{*}{\textbf{LLM}} & \multirow{2}{*}{\textbf{MMVP}} & \multirow{2}{*}{\textbf{RWQA}} & \multicolumn{2}{c}{\textbf{CV-Bench}} \\
 & & &   & \textbf{2D} & \textbf{3D} \\
 \midrule
 
GPT-4V & - &50.0 &61.4 &64.3 &73.8 \\
Cambrian-1 & Llama 3-8B& 51.3 &  64.2 & 72.3 & 72.0\\
\midrule

Cambrian-dev & Vicuna-7B & 30.0 & 54.0 & 55.5 & 53.6\\
\newshortname{} & Vicuna-7B  & 24.7 & 54.8 & 56.6 & 59.5 \\
\rowcolor{cyan!20}
 \ourshortname{} & Vicuna-7B & 30.0 & \underline{56.9} & 58.4 & \underline{61.9} \\
\newshortname{} & Llama 3-8B & \underline{32.0} & 55.6 & \underline{59.4} &  \underline{61.9} \\
 \rowcolor{cyan!20}
 \ourshortname{} & Llama 3-8B & \textbf{34.7} & \textbf{57.3}& \textbf{60.6} & \textbf{68.9}\\
\bottomrule
\end{tabular}
}
\caption{Comparison with SoTA methods on benchmarks on \textbf{vision-centric benchmarks.}}
\label{tab:results2}
\vspace{-3mm}
\end{table}


\subsection{Ablation Analysis}\label{sec_ablation}

In this section, we conduct ablation experiments to thoroughly explore the impact of different components of our model on our task. Our ablations includes the Panther-VE modules, different visual encoders/LLMs, Panther-Bridge and the discussion on the comparison with fine-tuning. Note that most of our ablations are performed on both OpenLlama-3B (denoted as Llama-3B)~\cite{openlm2023openllama} and Vicuna-7B~\cite{vicuna}.

\paragraph{Modules in Panther-VE:} 
As illustrated in Table \ref{tab:prompts}, the integration of SP into the baseline results in a certain degree of improvement, with even more significant enhancement observed upon the additional incorporation of IP. We also conducted experiments with various text encoders, including the BGE-base (BGE-b) \cite{bge_embedding}, but observed lower performance compared to the CLIP text encoder. We hypothesize that the superior alignment between CLIP's image and text encoders contributes to its better performance.


\begin{table}[htbp]
\centering
\scalebox{0.85}{
\begin{tabular}{l | c c c c c c }
\toprule
\textbf{LLM}  & \textbf{VQA$^{\textnormal{v2}}$} & \textbf{GQA}  & \textbf{VQA$^{\textnormal{T}}$} & \textbf{MMVP} \\

\midrule
Llama-3B & 76.8 & 60.9 &	47.3 &	11.3  \\
w/  SP & 78.0 & 61.8 & 48.8 & 13.3 \\
\rowcolor{gray!20}
w/  SP +  IP (CLIP) & \textbf{79.4} & \textbf{64.2} & \underline{51.2} & \textbf{22.7} \\
w/  SP +  IP (BGE-b) & \underline{79.0} & \underline{63.8} & \textbf{52.2} & \underline{15.3} \\
\midrule
Vicuna-7B & 78.5 & 	62.0 &  58.2 & 24.7 \\
w/  SP & 79.8 & 63.4 & 59.7 &	27.0 \\
\rowcolor{gray!20}
w/  SP +  IP (CLIP) & \textbf{80.8} & 	\underline{65.2} &	\textbf{59.6}	& \textbf{30.0} \\
w/  SP +  IP (BGE-b) & \underline{80.7} & \textbf{65.6} & \underline{58.8} & \underline{28.0}\\
\bottomrule
\end{tabular}
}
\vspace{-1mm}
\caption{\textbf{Ablation on Panther-VE modules}, the vision encoder is CLIP. \textbf{IP}: Instruction-aware visual Prompt, \textbf{SP}: Shared Prompt,}
\label{tab:prompts}
\vspace{-3mm}
\end{table}

\paragraph{Ablation on Visual Encoders and LLMs:}

As shown in Table \ref{tab:vis_encoder} and Table \ref{tab:LLM}, we perform ablations on both visual encoders (DINOv2~\cite{oquab2023dinov2}, CLIP~\cite{radford2021clip} and SigLIP~\cite{zhai2023sigmoid}) and LLMs (Llama-3B~\cite{openlm2023openllama}, Vicuna-7B~\cite{vicuna}, Mistral-7B~\cite{jiang2023mistral7b}, Llama 3-8B\cite{dubey2024llama3herdmodels}).
It is evident that Panther consistently enhances the performance of various visual encoders and LLMs across all the four benchmarks.


\begin{table}[htbp]
\centering
    \scalebox{0.75}{
\begin{tabular}{l l | c c c c c c c c }
\toprule
\textbf{VE (size)} & \textbf{LLM}  & \textbf{VQA$^{\textnormal{v2}}$} & \textbf{GQA}  & \textbf{VQA$^{\textnormal{T}}$} & \textbf{MMVP} \\
\midrule
CLIP (336$^2$) &  Llama-3B & 76.8 & 60.9 &	47.3 &	11.3  	\\
\rowcolor{gray!20}
\multicolumn{2}{c|}{+ ours}  & 79.4 & 64.2 & 51.2 & 22.7    \\
CLIP (336$^2$) &  Vicuna-7B &78.5 & 	62.0 & 58.2 & 24.7   \\
\rowcolor{gray!20}
\multicolumn{2}{c|}{+ ours} & 80.8 & 	65.2&	59.6	&30.0  \\

\midrule

DINO (336$^2$)&  Llama-3B & 73.7 & 	60.2 &  38.5 & 11.3   \\
\rowcolor{gray!20}
\multicolumn{2}{c|}{+ ours} &  75.9 & 62.7 &  39.2 & 14.0 \\
DINO (336$^2$)& Vicuna-7B & 76.5  & 	62.7  & 46.9 & 30.7   \\
\rowcolor{gray!20}
\multicolumn{2}{c|}{+ ours} &   78.2 & 	64.5  & 46.0 & 36.7  \\

\midrule

SigLIP (384$^2$) & Llama-3B &  78.6 & 	61.8&	52.1	&14.7 \\
\rowcolor{gray!20}
\multicolumn{2}{c|}{+ ours} & 80.3 & 	63.5 &	54.9 &	18.0  \\

SigLIP (384$^2$) & Vicuna-7B &  79.3 & 	62.3&	58.7	&26.0  \\
\rowcolor{gray!20}
\multicolumn{2}{c|}{+ ours}  &   81.0 & 	65.0&	 60.4	&37.3 \\

\bottomrule
\end{tabular}
}
\caption{\textbf{Ablation on vision encoders (VE). }}
\label{tab:vis_encoder}
\end{table}




\begin{table}[htbp]
    \centering
        \scalebox{0.8}{
\begin{tabular}{l | c c c c}
\toprule
\textbf{LLM}  & \textbf{VQA$^{\textnormal{v2}}$} & \textbf{GQA}  & \textbf{VQA$^{\textnormal{T}}$} & \textbf{MMVP} \\
\midrule
Llama-3B & 76.8 & 60.9 &	47.3 &	11.3    \\
\rowcolor{gray!20} + ours &  79.4 & 64.2 & 51.2 & 22.7     \\
Vicuna-7B &78.5 & 	62.0 &  58.2 & 24.7  \\
\rowcolor{gray!20}
+ ours & 80.8 & 	65.2&		59.6	&30.0   \\
Mistral-7B & 77.7 & 61.0 &	 54.7 & 18.0	 \\
\rowcolor{gray!20} + ours & 80.4 & 	64.9 &	 55.1 &    25.3 \\
Llama3-8B & 79.7 & 	63.3 &	 58.4&   32.0 \\
\rowcolor{gray!20} + ours & 81.2 & 65.5 &  59.5& 34.7    \\
\bottomrule
\end{tabular}
}
\caption{\textbf{Ablation on LLMs}, based on CLIP vision encoder.}
\label{tab:LLM}
\end{table}

\paragraph{Token-Pruning Analysis:}
In this section, we present an analysis of token pruning results in Table \ref{token_prune}, encompassing both the training time-cost (hours, h) and the corresponding evaluation outcomes. Our findings indicate that a threshold of $\tau=0.95$ achieves the optimal balance between training time and final performance.


\begin{table}[htbp]
\centering
\scalebox{0.75}{
\begin{tabular}{l | c c c c |c}
\toprule

\textbf{LLM}  & \textbf{VQA$^{\textnormal{v2}}$} & \textbf{GQA}  & \textbf{VQA$^{\textnormal{T}}$} & \textbf{MMVP} & \textbf{Cost (h)}\\

\midrule
Llama-3B & 76.8 & 60.9 &	47.3 &	11.3  & $\sim$ 5.0 \\
+ ours, $\tau=1.00$ & \textbf{79.5} & \textbf{64.4} & \textbf{51.2} & \textbf{23.3}  & $\sim$ 10.7\\
+ ours, $\tau=0.97$ & {79.3} & \textbf{64.4} & {51.0} & {21.7}  & $\sim$ 8.3\\
\rowcolor{gray!20}
+ ours, $\tau=0.95$ & \underline{79.4} & {64.2} & \textbf{51.2} & \underline{22.7}  & $\sim$ 6.2\\
+ ours, $\tau=0.90$ & {79.2} & {63.8} & {50.8} & {22.0}  & $\sim$ 5.5\\

\midrule
Vicuna-7B & 78.5 & 	62.0 &  58.2 & 24.7 & $\sim$ 9.5\\
+ ours, $\tau=1.00$  & \textbf{81.1} & 	\textbf{65.3} &	\textbf{60.1}	&\textbf{32.3} & $\sim$ 23.5\\
+ ours, $\tau=0.97$  & \textbf{81.1} & 	{65.0} &	\underline{59.9}	&\underline{30.3} & $\sim$ 18.0\\
\rowcolor{gray!20}
+ ours, $\tau=0.95$  & {80.8} & 	\underline{65.2} &	59.6	&{30.0} & $\sim$ 12.8\\
+ ours, $\tau=0.90$  & {80.5} & 	{64.9} &	59.2	&{29.3} & $\sim$ 11.1\\
\bottomrule
\end{tabular}
}

\caption{\textbf{Token-pruning analysis}, based on CLIP, Llama-3B and Vicuna-7B. The $\tau$ is the cosine similarity threshold.}
\label{tab:token_prune}
\end{table}

\paragraph{Discussion on Fine-Tuning:}
We also conduct experiments using full fine-tuning (FT) on visual encoder (VE) backbone and combine it with our approach. The results are detailed in Appendix with main findings that: FT on VE can improve the performance, while our IP can further improve performance based on FT, demonstrating the effectiveness of our approach.

\begin{figure}[htbp]
\centering
\includegraphics[width=1.\linewidth]{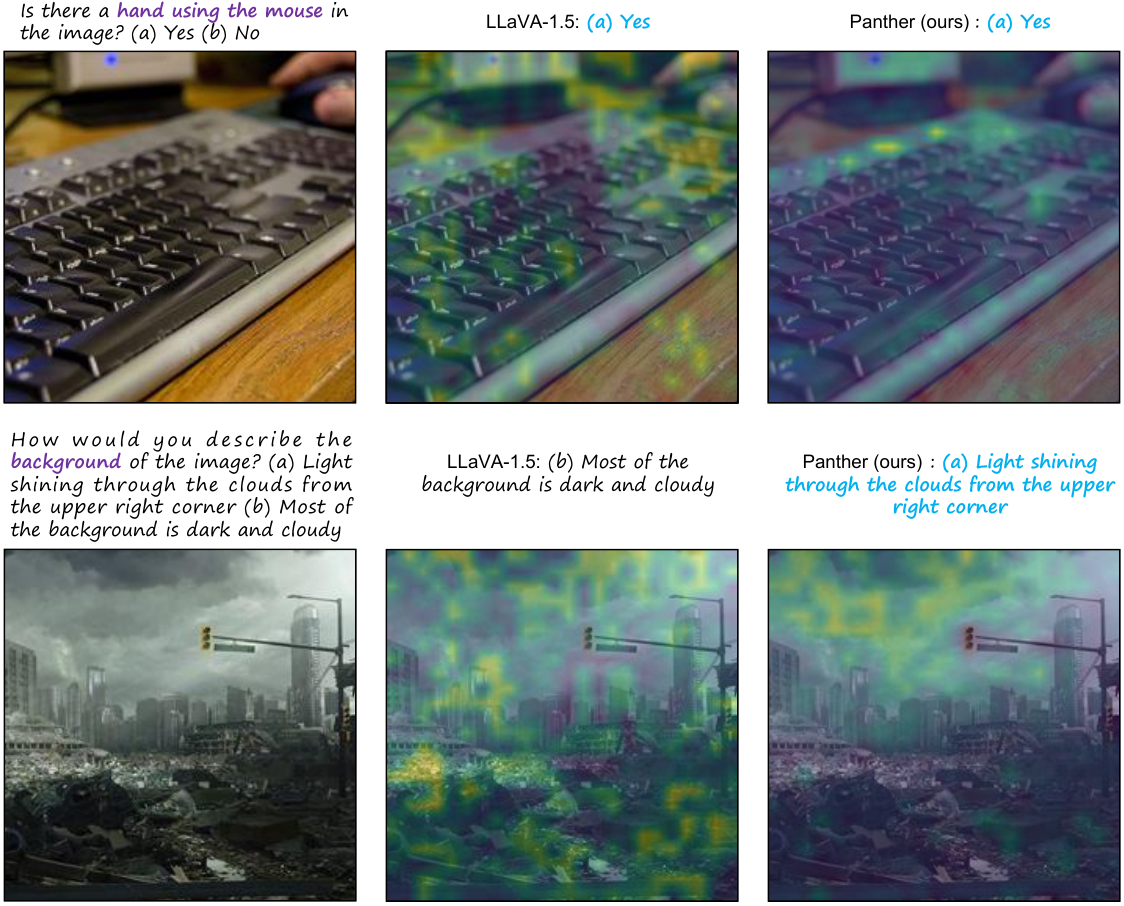} \\
\caption{ Qualitative comparisons for representative scenarios. The purple represents the key objects in instruction user sought, while the blue denotes the correct answer.}
\label{fig_quality_eval}
\end{figure}

\subsection{Qualitative Evaluation}
Here we provide qualitative evaluation of our MLLM as depicted in Figure  \ref{fig_quality_eval}: 

\noindent \textbf{First row: Keyboard} image with instruction “Is there a hand using the mouse in the image?".

\underline{\textit{LLaVA-1.5}}: Correctly answers “Yes”, but the attention is spread out, lacking focus on the hand and mouse.

\underline{\textit{Panther}}: Also answers “Yes”, with attention clearly concentrated on the hand and mouse area, indicating a better alignment with the question’s intent.

\noindent \textbf{Second row: Cloudy Cityscape} image with instruction “How would you describe the background of the image?".

\underline{\textit{LLaVA-1.5}}: Incorrectly selects (b), “dark and cloudy”, with a dispersed attention map, missing the light in the upper right corner.

\underline{\textit{Panther}}: Correctly selects (a), describing "light shining through the clouds", with attention concentrated in the upper right, capturing the subtle detail requested.

Clearly, our Panther demonstrates superior focus on instruction-relevant areas, effectively addressing the \textit{Amblyopia} issue.

%% file: sec/X_suppl.tex
\maketitlesupplementary

\paragraph{Discussion on Fine-Tuning:}
We have conducted a series of experiments using full fine-tuning (FT) for comparative analysis, thereby validating the effectiveness of our approach. As depicted in Table \ref{tab:ft}, the straightforward full fine-tuning (FT) yields a modest improvement in the performance of the vanilla \newshortname{} model, akin to the enhancement achieved by the Shared Prompt (SP) method. After combined with our Instruction-aware visual Prompt (IP), both the FT and SP can be further improved, demonstrating the exceptional capabilities and contributions of our method.

\begin{table}[htbp]
\centering
\scalebox{0.85}{
\begin{tabular}{l | c c c c c c c c }
\toprule
\textbf{LLM}  & \textbf{VQA$^{\textnormal{v2}}$} & \textbf{GQA}  & \textbf{VQA$^{\textnormal{T}}$} & \textbf{MMVP} \\
\midrule
Llama-3B & 76.8 & 60.9 &	47.3 &	11.3   \\
w/ FT & 78.7 & 62.7 &	48.0 &	12.0  \\
w/ SP & 78.0 & 61.8 & 48.8 & 13.3 \\
\rowcolor{gray!20}
w/ SP + IP & \underline{79.4} & \textbf{64.2} & \underline{51.2} & \textbf{22.7}  \\
\rowcolor{gray!20}

w/ FT + IP  & \textbf{79.7} & \underline{63.7} & \textbf{51.5} & \underline{18.6}   \\
\midrule
Vicuna-7B & 78.5 & 	62.0 &  58.2 & 24.7 \\
w/ FT  & 80.7 & 63.7 &  59.5 & 29.3  \\
w/ SP & 79.8 & 63.4 & \underline{59.7} &	27.0 \\
\rowcolor{gray!20}
w/ SP + IP & \underline{80.8} & 	\underline{65.2} &	59.6	&\underline{30.0} \\
\rowcolor{gray!20}
w/ FT + IP & \textbf{81.2} & \textbf{65.5} &  \textbf{60.2} & \textbf{34.0}  \\
\bottomrule
\end{tabular}
}
\caption{\textbf{Discussion on Fine-tuning}, based on CLIP, Llama-3B and Vicuna-7B. \textbf{FT}: Fine-Tuning, \textbf{SP}: Shared Prompt, \textbf{IP}: Instructed-aware visual Prompt.}
\label{tab:ft}
\end{table}

\paragraph{Panther-Bridge token-pruning on multi-turns: }

Denoting multi-turns ($\ge 3$) visual token embeddings as $\{\mathbf{T}^{\prime_0}, \mathbf{T}^{\prime_1}, \mathbf{T}^{\prime_2}, ... , \mathbf{T}^{\prime_{K-1}} \}$. \\
\noindent \textbf{Step 1}: We keep the $\mathbf{T}^{\prime_0}$ and prune the remaining turns' tokens $\mathbf{T}^{\prime_k} (1\le k \le K-1)$ if the cosine similarity exceeds a threshold $\tau$ and retain the useful tokens as:
\begin{equation}
\mathbf{T}_{\text{useful-1}}^{\prime_k} = \{ T_{i}^{\prime_k} \mid \cos(T_{i}^{\prime_0}, T_{i}^{\prime_k}) \leq \tau \}.
\label{eq:cos_prune}
\end{equation}
\noindent \textbf{Step 2}: We keep the $\mathbf{T}^{\prime_0}, \mathbf{T}_{\text{useful-1}}^{\prime_1}$ and prune the remaining turns' tokens $T_{\text{useful-1}}^{\prime_k} (2\le k \le K-1)$:
\begin{equation}
\mathbf{T}_{\text{useful-2}}^{\prime_k} = \{ T_{\text{useful-1},i}^{\prime_k} \mid \cos(T_{\text{useful-1},i}^{\prime_1}, T_{\text{useful-1},i}^{\prime_k}) \leq \tau \}.
\label{eq:cos_prune}
\end{equation}
Here only token with same spatial index need to be pruned.

\noindent \textbf{Step n}: Above process can be looped until the last turn, thus we omit the equation here but provided a PyTorch-style pseudo code for the whole process in Algorithm \ref{alg_0}.

\begin{algorithm}[tb]
   \caption{PyTorch pseudocode for multi-turn visual token embeddings.}
   \label{algo:DINO}
    \definecolor{codeblue}{rgb}{0.25,0.5,0.5}
    \lstset{
      basicstyle=\fontsize{7.2pt}{7.2pt}\ttfamily\bfseries,
      commentstyle=\fontsize{7.2pt}{7.2pt}\color{codeblue},
      keywordstyle=\fontsize{7.2pt}{7.2pt},
    }
\begin{lstlisting}[language=python]
# T_v: visual tokens tensor with shape (K, n, d1)
# K: overall turns
# n: token numbers for each turn 
# d1: token embedding dim size 
# cos, tau: cosine similarity and the threshold
# T_r: output useful visual tokens for each turn
def main():
    # keep the first turn T_v[0]
    T_u = [{'idx_list': range(n), 'tensor': T_v[0]}]
    # Step 1: Iterate over each turn from k = 1 to K-1
    # and prepare index
    ref_T = T_u[0]
    for k in range(1, K):
        cur_T = {'idx_list': range(n), 'tensor': T_v[k]}
        useful_T = prune_tokens(cur_T, ref_T)
        T_u.append(useful_T)
    
    # Step 2~n: recursive pruning 
    for k in range(2, K):
        new_T = T_u[k:] # these tokens are further pruned
        T_u = prune_concat(T_u[:k], new_T)

    # delete index and return tokens tensor
    T_r = []
    for T_u_k in T_u:
        T_r.append(T_u_k['tensor'])
    return T_r
    
def prune_concat(useful_T_all, new_T):
    ref_T = useful_T_all[-1]
    for k in range(len(new_T)):
        cur_T = {'idx_list': range(n), 'tensor': new_T[k]}
        useful_T = prune_tokens(cur_T, ref_T)
        useful_T_all.append(useful_T)
    return useful_T_all
    
def prune_tokens(cur_T, ref_T):
    """
    Args:
        cur_T: current tokens to be pruned 
        cur_T = {'idx_list:':[...], 'tensor':[...]}.
          e.g. idx_list: [10,15,16,17,20]
        
        ref_T: reference tokens
        ref_T = {'idx_list':[...], 'tensor':[...]}.
          e.g. idx_list: [10,14,16,17,22]
      pruning only on same indices [10,16,17], 
      while keeping [15,20] 
    """
    ret_T = ['idx_list':[], 'tensor':[]]  
    # Store retained tokens
    for i, spatial_i in enumerate(cur_T['idx_list']):
        cur_token = cur_T['tensor'][i]
        if spatial_i not in ref_T['idx_list']: 
            ret_T['idx_list'].append(spatial_i)
            ret_T['tensor'].append(cur_token)
        else: # match the spatial index
            j = torch.where(ref_T['idx_list']==spatial_i)
            ref_token = cur_T['tensor'][j]
            cosine_sim = cos(
                cur_token, ref_token, dim=-1
                )
            if cosine_sim <= tau:
                ret_T['idx_list'].append(spatial_i)
                ret_T['tensor'].append(cur_token)
            else:
                pass # pruned
    return ret_T
\end{lstlisting}
\label{alg_0}
\end{algorithm}

\clearpage
\paragraph{Survey on Visual Token-Pruning/Merging:}
Filtering out less informative visual tokens while preserving essential ones is critical for improving the efficiency of Transformer-based models such as ViT and MLLMs. Various approaches address this challenge:

In ViT: Several works, such as SPViT and EViT \cite{spvit,evit}, propose combining pruned tokens into a single representation.
TokenLearner \cite{tokenlearner} employs an MLP to reduce the number of tokens.
LIT \cite{pan2022less} introduces deformable token-merging layers for pooling between stages.
Methods like Token-Pooling \cite{tokenpooling} and Token Merging \cite{bolya2022token} aggregate and merge similar tokens to condense token representations effectively.

In MLLMs:
Visual tokens dominate the token space in these models, with instances like LLaVA-1.5 \cite{llava} utilizing 576 tokens, and even higher numbers required for high-resolution images \cite{liu2024llava16,liu2024improved,dong2024internlm,li2024monkey,wang2024qwen2vl,yao2024minicpmv} or video understanding \cite{wang2024tarsier,lin2023video,xu2024pllava,zhang2023video}.
Q-Former \cite{dai2023instructblip} and Resampler \cite{alayrac2022flamingo} leverage cross-attention mechanisms to condense any number of visual tokens into a fixed-length representation.
Token-Packer \cite{li2024tokenpacker} enriches condensed visual tokens through low-resolution point queries and fine-grained region-to-point injection.
Neighbor-Aware Token Pruning \cite{zhang2024treat} utilizes selective token attention, inactive head pruning, and layer dropping to streamline token processing.
VLoRA \cite{ma2024visual} uniquely converts visual tokens into LLM weights, providing a novel approach to token efficiency.
In video understanding, AURORACAP \cite{chai2024auroracap} implement the token merging \cite{bolya2022token} strategy, reducing the number of input visual tokens to address the overhead caused by
lengthy video sequences. For long video understanding efficiency, MA-LMM \cite{he2024ma} merge tokens between similar frames and \cite{zhang2024longcontexttransferlanguage} dynamically removes less important tokens layer by layer to reduce computation while preserving critical information.
These advancements significantly enhance the efficiency of ViT and MLLMs by reducing computational overhead while maintaining robust representation quality.